\pdfoutput=1

\documentclass[11pt]{article}

\usepackage[]{acl}

\usepackage{times}
\usepackage{latexsym}
\usepackage{arydshln}
\usepackage[T1]{fontenc}

\usepackage[utf8]{inputenc}

\usepackage{microtype}
\usepackage{booktabs}
\usepackage{multicol}
\usepackage[utf8]{inputenc}
\usepackage{mathtools}
\usepackage{amsmath}
\usepackage{color,soul}
\usepackage[normalem]{ulem}
\newcommand{\intitle}[1]{\vspace{6pt}\noindent {\bf {#1}.}}
\usepackage{geometry}
\usepackage{authblk}

\newcommand{\lea}[1]{}
\newcommand{\leacom}[1]{}
\newcommand{\least}[1]{}
\newcommand{\oa}[1]{}
\newcommand{\gabis}[1]{}
\newcommand{\gabisrep}[2]{}
\newcommand{\gabist}[1]{}
\newcommand{\ub}[1]{}
\newcommand{\ubst}[1]{}

\title{A Computational Acquisition Model for Multimodal Word Categorization}

\author[1,2]{\bf{Uri Berger}}
\author[1]{\bf{Gabriel Stanovsky}}
\author[1]{\bf{Omri Abend}}
\author[2]{\bf{Lea Frermann}}
\affil[1]{School of Computer Science and Engineering, The Hebrew University of Jerusalem}
\affil[2]{School of Computing and Information Systems, University of Melbourne}
\affil[ ]{\texttt{\{uri.berger2, gabriel.stanovsky, omri.abend\}@mail.huji.ac.il}}
\affil[ ]{\texttt{lea.frermann@unimelb.edu.au}}

\begin{document}

\maketitle
\begin{abstract}
Recent advances in self-supervised modeling of text and images open new opportunities for computational models of child language acquisition, which is believed to rely heavily on cross-modal signals. However, prior studies have been limited by their reliance on vision models trained on large image datasets annotated with a pre-defined set of depicted object categories. This is (a) not faithful to the information children receive
and (b) prohibits the evaluation of such models with respect to category learning tasks, due to the pre-imposed category structure. We address this gap, and present a cognitively-inspired, multimodal acquisition model, trained from image-caption pairs on naturalistic data using cross-modal self-supervision. We show that the model learns word categories and object recognition abilities, and presents trends reminiscent of those reported in the developmental literature. We make our code and trained models public for future reference and use\footnote{\href{https://github.com/SLAB-NLP/multimodal_clustering}{github.com/SLAB-NLP/multimodal\_clustering}}.
\end{abstract}

\section{Introduction} \label{introduction}

To date, the mechanisms underlying the efficiency with which infants learn to speak and understand natural language remain an open research question.
Research suggests that children leverage contextual, inter-personal and non-linguistic information.
Visual input is a case in point: when spoken to, infants visually perceive their environment, and paired with the input speech, the visual environment could help bootstrap linguistic knowledge \cite{tomasello_1996}. Unlike social cues, visual input has a natural physical representation, in the form of pixel maps or videos.

Previous multimodal language acquisition studies either considered toy scenarios with small vocabularies~\cite{roy2002learning,frank2007bayesian}, or used visual encoders that were pretrained on large labeled data bases such as ImageNet \cite{deng_et_al_2009} or Visual Genome \cite{krishna_et_al_2017}. This has two drawbacks: first, systematic access to labeled data is a cognitively implausible assumption in a language acquisition setting; second, imposing a pre-defined categorization system precludes studying categories that emerge when learning from unlabeled multimodal data. This type of setting more closely resembles the data underlying early language learning at a time when the child has only acquired little conceptual information. Although the subject of much psycholinguistic work, the computational study of multimodal word categories, formed without recourse to manual supervision has  been scarcely addressed in previous work.

\begin{figure}[tb]
    \centering
    \includegraphics[width=7cm]{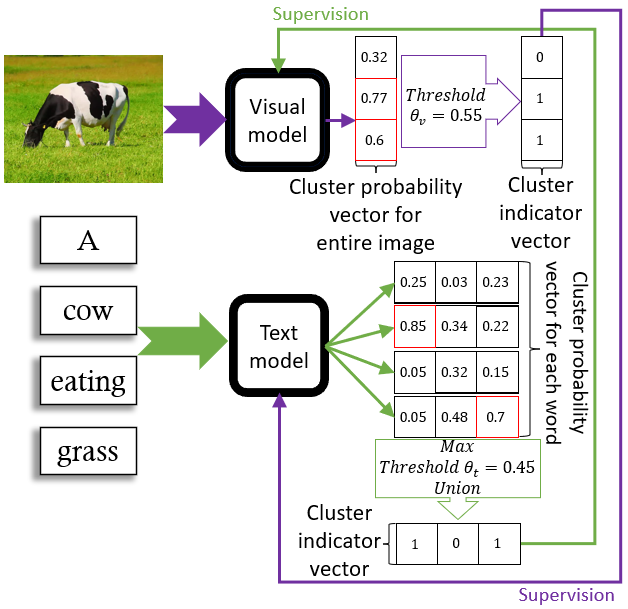}
    \caption{Model overview. Given an image-caption pair (left), both the visual (top) and textual encoder (bottom) generate a binary vector indicative of the clusters associated with the current input. The text/visual model predicts a probability vector over clusters per word/image. Vectors are mapped to a binary space using thresholding. The modality-specific vectors provide mutual supervision during training (right).
    }
    \label{fig:multimodal_model_desc}
\end{figure}

We present a model that learns categories as clusters of words from large-scale, naturalistic multi-modal data without any pre-training. Given (image, caption)-pairs as input, the model is trained to cluster words and images in a shared, latent space, where each cluster pertains to a semantic category.
In a self-supervised setup, a neural image classifier and a co-occurrence based word clustering module provide mutual supervision through 
joint training with an expectation-maximization (EM) style algorithm. Figure~\ref{fig:multimodal_model_desc} illustrates our model.

Our input representation is cognitively plausible in that we use raw image data (pixels), and train our model on a comparatively small data set (Table~\ref{table:models_statistics}). However, we follow previous work~\cite{kadar_et_al_2015, nikolaus_fourtassi_2021} in using written and word-segmented text input. Young infants do not have access to such structure, and this assumption therefore deviates from cognitive plausibility (but opens avenues for future work).

\begin{table}
\setlength{\tabcolsep}{0.15em} 
\centering
\begin{footnotesize}
\begin{tabular}{lcccr} 
\toprule
Model & parameters & token & (img, capt) & train time \\
\midrule
BERT & 110M & 3300M & -- & 64 TPU days \\ 
word2vec & 1B & 100B & -- & 60 CPU days \\
CLIP & 63M & --& 400M &  10K GPU days\\ 
\bf{Ours} & \bf{3M} & \bf{3M} & \bf{280K} & \bf{4 GPU days} \\ 
Children & -- & 13K/day & -- & -- \\
\bottomrule
\end{tabular}
\end{footnotesize}
\caption{Comparing the size, training data and training time of popular self-supervised models (BERT$_{\text{base}}$, word2vec (300d, google-news) and CLIP with ResNet50X64) against our model (Ours, bold) and typical child input \citep[Children;][]{gilkerson_2017}. For the multimodal models (CLIP and Ours) we only mention the size of the text encoder.}
\label{table:models_statistics}
\end{table}

We show that semantic and visual knowledge emerges when training on the self-supervised categorization task. In a zero-shot setup, we evaluate our model on (1)~word concreteness prediction as a proxy for noun identification, one of the first cues for syntax in infant language acquisition \cite{fisher_et_al_1994}; (2) visual classification and object segmentation. We also study the emerging latent word clusters and show that words are clustered syntagmatically~\cite{Saussure1916}: words representing entities that are likely to occur together are more likely clustered together (e.g., dog-bark), than words that share taxonomic categories (e.g., dog-cat). This concurs with findings that young children acquire syntagmatic categories more readily than taxonomic categories~\cite{sloutsky_2017}.


\section{Background and Related Work} \label{background}
We briefly review previous studies of unimodal learning without pretraining (for both text and vision) and multimodal learning (studied mainly in acquisition implausible settings) to highlight the gap that this work addresses.

\intitle{Unimodal Learning} 
Self-supervised learning without pre-training has been extensively studied, but predominantly in a unimodal scenario or under cognitively implausible assumptions.
For the text modality, large language models have been developed in recent years \citep[e.g., BERT;][]{devlin_et_al_2019}, trained on large unlabeled text corpora (Table~\ref{table:models_statistics}).
A more cognitively motivated model is BabyBERTa \cite{huebner_et_al_2021},  a smaller version of RoBERTa \cite{liu_et_al_2019} (also) trained on transcribed child directed speech.
In the visual domain, self-supervision is typically implemented as contrastive learning, training the model to align corrupted images with their original counterparts
~\citep{chen_et_al_2020}, with subsequent fine-tuning.

\intitle{Multimodal Language Learning}
Early language acquisition studies \cite{roy2002learning} considered toy scenarios with small vocabularies and used heuristics for image processing.
\citet{silberer_lapata_2014} model multi-modal human categorization using human-annotated feature vectors as input for a multimodal self-supervised autoencoder, while we learn the features from raw images. 

Unlike our work, recent work on cross-modal language learning \cite{kadar_et_al_2015, chrupala-etal-2017-representations, ororbia_et_al_2019, nikolaus_fourtassi_2021} typically use Convolutional Neural Networks, pre-trained on large labeled data bases like ImageNet~\cite{deng_et_al_2009}, 
or alternatively \citep[e.g.,][]{lu_et_al_2019,chen_et_al_uniter_2020} use object detectors pre-trained on Visual Genome \cite{krishna_et_al_2017} as the visual model. 

Few studies assume no prior knowledge of the grounded modality. Most related to our study is CLIP~\cite{radford_et_al_2021}, 
a pre-trained off-the-shelf model trained to project matching images and captions to similar vectors. CLIP assumes a multimodal joint space which is continuous, unlike our binary space.
\citet{liu_et_al_2021} use CLIP pretrained encoders to learn cross-modal representations with a similar training objective as ours. They discretize the output of the encoders by mapping it to the closest vector from a finite set of learned vectors, which can be viewed as a form of categorization. CLIP-based works are trained to match entire sentences to images and have no explicit representation of words and phrases. We therefore view it as a cognitively less plausible setting than is presented in the current study. Nevertheless, we include CLIP as a point of comparison when applicable.

\section{Model} \label{methods}

Our goal is to learn the meaning of words and raw images through mutual supervision by mapping both modalities to a joint representation. Intuitively, given a visual input paired with relevant text (approximating a typical learning scenario), the output for each of the modalities is a binary vector in $\{0,1\}^N$, with non-zero dimensions indicating the clusters\footnote{We use the term \emph{clusters} for the output of models and the term \emph{categories} for ground truth classification of elements (e.g., human defined categories).} to which the input is assigned and $N$ is the total number of clusters (a predefined hyper-parameter).
The clusters are unknown a priori and are formed during training.
The goal is to assign matching text and image to the same clusters.
In order to minimize assumptions on innate knowledge or pre-imposed categories available to the language learner, and enable the study of emerging categories from large-scale multi-modal input data, we deliberately avoid any pre-training of our models.  

\subsection{Visual Encoder} 
The visual encoder (Figure~\ref{fig:multimodal_model_desc}, top) is a randomly initialized ResNet50 \cite{he_et_al_2016}, without pretraining. We set the output size of the network to $N$ (the number of clusters) and add an element-wise sigmoid layer.\footnote{This is a multi-label categorization task (i.e., an image can be assigned to multiple clusters): Cluster assignments do not compete with one another, which is why we chose sigmoid over softmax.} To produce the binary output and predict the clusters given an input image, we apply a hyper-parameter threshold $\theta_v$ to the output of the sigmoid layer.

\subsection{Text Encoder} \label{textual_encoder}
The text encoder (Figure~\ref{fig:multimodal_model_desc}, bottom) is a simple probabilistic model based on word-cluster co-occurrence, which is intuitively interpretable and makes minimal structural assumptions.
Given a sentence, the model assigns each word to at most one cluster. 
The sentence is assigned to the union of the clusters to which the words in it are assigned. Formally, given a sentence $s{=}(w_1, w_2, ..., w_n)$ of words $w_i$, and an assignment of the words to clusters $f{:}\{w_1,...,w_n\}{\rightarrow}\{1,...,N\} \cup \{ \emptyset \}$, the clusters to which the sentence is assigned are:
$
    \bigl\{ c | \text{if } \exists w_i  \text{ s.t. } f(w_i){=}c \bigr\}_{c=1}^N
$. 
When assigning words to clusters, we make two simplifying assumptions: (1) the probability that a word is assigned to a specific cluster is independent of the linguistic context, meaning that we assign to clusters on the type- rather than the token level (a reasonable assumption given that children learn single words first); (2) a single word cannot be assigned to more than one cluster,
{\it but it might be assigned to no cluster at all if it does not have a visual correspondent in the image} (e.g., function words).
Under these assumptions, the encoder estimates $P(c|w)$ for each $c \in \{1,\dots,N\}$ and for each $w \in V$, where $V$ is the vocabulary.
If the probability of assigning a given word in a sentence to any of the clusters exceeds a hyper-parameter threshold $\theta_t$, it is assigned to the cluster with the highest probability, otherwise it is not assigned to any cluster.
Formally:

\begin{equation*}
    f(w) =
    \begin{cases}
        \underset{c \in [N]}{\text{argmax}} \, P(c|w) & \text{if } \underset{c \in [N]}{\text{max}} \, P(c|w) \geq \theta_t\\
        \emptyset & \text{else}
    \end{cases}
\end{equation*}
In the next step, we define the word-cluster associations $P(c|w)$. We estimate these using Bayes Rule,
\begin{equation}
    P(c|w) = \frac{P(w|c) P(c)}{P(w)}
    \label{eqn:bayes}
\end{equation}
$P(w|c)$ is defined as the fraction of all predictions of cluster $c$ from the visual encoder, in which $w$ occurred in the corresponding caption.
We instantiate the prior cluster probability $P(c)$ as uniform over all clusters.\footnote{We also tried instantiating $P(c)$ as the empirical cluster distribution as predicted by the visual encoder. However, the noisy initial predictions lead to a positive feedback loop leading to most clusters being unused.}

Finally, for a given word $w$, we estimate $P(w){=}\sum_{i=1}^{N} P(c_i) P(w|c_i)$. 
Intuitively, we expect that a concrete word would repeatedly occur with similar visual features (of the object described by that word), therefore repeatedly co-occurring with the same cluster and receiving a high assignment probability with that cluster, whereas abstract words would co-occur with multiple clusters, therefore not being assigned to any cluster.

\section{Training} \label{training_algorithm}
At each training step, the model observes a batch of (image, caption)-pairs. We first perform inference with both encoders, and then use the results of each encoder's inference to supervise the other encoder.

\intitle{Text Encoder} Given the list of clusters predicted by the visual encoder and a tokenized caption $s{=}\{w_1, ..., w_n\}$, for each $w_i{\in}s$ and for each cluster $c_j$ predicted by the visual encoder, we increment $count(c_j)$ and $count(w_i, c_j)$. These are needed to compute the probabilities in equation~(\ref{eqn:bayes}).

\intitle{Visual Encoder} For each input image and corresponding cluster vector predicted by the text encoder, we use binary cross entropy loss comparing the output of the sigmoid layer with the predicted cluster vector and use Backpropagation to update the parameters of the ResNet model.

\section{Experiments} \label{experiments}

We trained our model on the 2014 split of MSCOCO~\cite{lin_et_al_2014}, a dataset of naturalistic images with one or more corresponding captions, where each image is labeled with a list of object classes it depicts. MSCOCO has 80 object classes, 123K images and 616K captions (split into 67\% train, 33\% test).
We filtered out images that did not contain any labeled objects, 
and images that contained objects with a multi-token label (e.g., ``fire hydrant'').\footnote{This  was enforced because multi-token labels do not necessarily map to a single cluster, rendering the evaluation more difficult. We plan to address this gap in future work.}
After filtering, we are left with 65 ground-truth classes. The filtered training (test) set contains 56K (27K) images and 279K (137K) captions. We set apart 20\% of the training set for hyper-parameter tuning.

We trained our model with a batch size of 50 until we observed no improvement in the F-score measure from Section \ref{semantic_word_clustering} (40 epochs). Training took 4 days on a single GM204GL GPU. We used $N{=}150$ clusters, $\theta_t{=}0.08$, and  $\theta_v{=}0.5$. The visual threshold $\theta_v$ was first set heuristically to $0.5$ to avoid degenerate solutions (images being assigned to all or no clusters initially). Then, $N$ and $\theta_t$ were determined in a grid search, optimizing the F-score measure from Section \ref{semantic_word_clustering}.
We used spaCy \cite{honnibal2017spacy} for tokenization.

\subsection{Semantic Word Categorization} \label{semantic_word_clustering}

\intitle{Background}
Semantic word categorization is the clustering of words based on their semantic features.
Psycholinguistic studies have shown that children use semantic word categories to solve linguistic tasks by the end of the second year of life \citep[e.g.,][]{styles_plunkett_2009}.
There is a long established fundamental distinction between {\it syntagmatic} (words that are likely to co-occur in the same context) and {\it paradigmatic} relations (words that can substitute one another in a context without affecting its grammaticality or acceptability) ~\cite{Saussure1916}. Each relation type invokes a different type of word categories (syntagmatic relations invoke syntagmatic categories, or {\it associative} categories; paradigmatic relation invoke {\it taxonomic} categories).
Despite an acknowledgement that infants, unlike adults, categorize based on syntagmatic criteria more readily than on paradigmatic criteria \citep[``The Syntagmatic-Paradigmatic shift'';][]{ervin1961changes}, and empirical evidence that syntagmatic categories might be more important for word learning than taxonomic categories \cite{sloutsky_2017}, computational categorization studies and datasets predominantly focused only on taxonomic hierarchies~\cite{silberer_lapata_2014,frermann_lapata_2016}.

\intitle{Setting}
Our model's induced clusters are created by using the text encoder to predict, for each word, the most likely cluster.

We evaluated induced clusters against a taxonomic and a syntagmatic reference data set. First, we followed \citet{silberer_lapata_2014}, used the categorization dataset from \citet{fountain_lapata_2010}, and transformed the dataset into hard categories by assigning each noun to its most typical category as extrapolated from human typicality ratings. The resulting dataset contains 516 words grouped into 41 taxononmic categories. We filtered the dataset to contain only words that occur in the MSCOCO training set and in the word2vec \cite{mikolov_et_al_2013} dictionary, obtaining the final dataset with 444 words grouped into 41 categories.

In order to quantify the syntagmatic nature of the induced clusters, we used a large dataset of human word associations, the "Small World of Words" \citep[SWOW,][]{de2019small}. SWOW was compiled by presenting a cue word to human participants and requesting them to respond with the first three words that came to mind. The association strength of a pair of words $(w_1, w_2)$ is determined by the number of participants who responded with $w_2$ to cue word $w_1$. Prior work has shown that word associations are to a large extent driven by {\it syntagmatic}  relations~\cite{santos_chaigneau_simmons_barsalou_2011}. 

\intitle{Comparison with other models}
We compare against several word embedding models,\footnote{We omit \citet{silberer_lapata_2014} since we were unable to obtain their model.} where for each model we first induce embeddings, which we then cluster  into K=41 clusters (the number of taxonomic gold classes) using K-Means.
We compare against a {\bf text-only} variant of our model\footnote{It was not clear how to design co-occurrence based vision-only baselines, as images do not naturally factorize into concepts/regions, unlike text which is divided into words.} by creating a co-occurrence matrix $C$ where $C_{i,j}$ is the number of captions in which tokens $i,j$ in the vocabulary co-occur. The normalized rows of $C$ are the vector embeddings of words in the vocabulary.
We compare against off-the-shelf {\bf word2vec} and $\text{\bf BERT}_\text{\bf BASE}$ embeddings. For BERT, given a word $w$, we feed an artificial context (``this is a $w$'') and take the embedding of the first subword of $w$. We also include the multi-modal {\bf CLIP}, using prompts as suggested in the original paper (``a photo of a $w$'').\footnote{We also tried BERT and CLIP feeding as input the target word only. Results for CLIP slightly decreased, while results for BERT decreased significantly.}
Finally, we include a randomized baseline, which assigns each word at random to one of 41 clusters. 
Implementation details can be found in Appendix~\ref{app:implementation}.

\intitle{Taxonomic categorization}
We use the F-score metric following \citet{silberer_lapata_2014}. The F-value of a (gold class, cluster)-pair is the harmonic mean of precision and recall defined as the size of intersection divided by the number of items in the cluster and the number of items in the class, respectively. The F-score of a class is the maximum F-value attained at any cluster, and the F-score of the entire clustering is the size-weighted sum of F-scores of all classes. We report performance over five random restarts for all models.

\begin{table}
\centering
\begin{tabular}{ll} 
\toprule
Model & F-Score \\
\midrule
Random & 0.15 $\pm$ 0.0032 \\\hdashline
Text-only & 0.26 $\pm$ 0.0098 \\
\bf{Word2vec} & \bf{0.40 $\pm$ 0.0172} \\
$\text{BERT}_\text{BASE}$ & 0.33 $\pm$ 0.011 \\\hdashline
CLIP & 0.38 $\pm$ 0.0142 \\
Ours & 0.33 $\pm$ 0.0109 \\
\bottomrule
\end{tabular}
\caption{Taxonomic categorization results, comparing a random baseline (top) against text-only (center) and multi-modal models (bottom), as mean F-score and std over 5 runs. Word2vec (bold) achieves the highest score.
}
\label{table:clustering_f_score}
\end{table}

Results are presented in Table \ref{table:clustering_f_score}. The text-only baseline improves results over a random categorization algorithm. Our multi-modal model grounded in visual input improves over its unimodal variant.
Our model is competitive with BERT and is surpassed by word2vec and CLIP. However, considering the small model and training data we used (see Table \ref{table:models_statistics}), our results are competitive.

\intitle{Syntagmatic categorization}
We quantify the syntagmatic nature of a clustering by the mean association strength (MAS) of pairs of words in the SWOW dataset, where association strength of a pair of words $(w_1, w_2)$ is again number of participants who responded with w2 to cue word w1. MAS is computed across all word pairs from the taxonomic dataset in which both words were assigned the same cluster by this clustering solution. 

\begin{table}
\centering
\begin{tabular}{ll} 
\toprule
Model & MAS \\
\midrule
Taxonomic & 5.72 \\
Random & 4.23 $\pm$ 1.88 \\\hdashline
Text-only & 5.47 $\pm$ 0.25 \\
Word2Vec & 6.65 $\pm$ 0.16 \\
BERT$_{\text{BASE}}$ & 5.75 $\pm$ 0.23 \\\hdashline
CLIP & 7.08 $\pm$ 0.41 \\
\bf{Ours} & \bf{7.45 $\pm$ 0.33} \\

\bottomrule
\end{tabular}
\caption{Sytagmatic categorization results on the same models as in Table~\ref{table:clustering_f_score}, reporting mean association strength (MAS) of pairs of clustered concepts in the SWOW dataset, over 5 random initializations. Taxonomic refers to the taxonomic gold categories. Our model (bold) achieves the highest score.}
\label{table:swow_mas}
\end{table}

Results are presented in Table \ref{table:swow_mas}. The multimodal models (ours and CLIP) outperform all unimodal models, an indication of the impact of multimodality on category learning: multimodal word learning shifts the learner towards syntagmatic relations more significantly than unimodal word learning. To our knowledge, this is the first computational result to support this hypothesis, shown empirically in human studies with infants
\cite{elbers1999lexical, mikolajczak2015associative}. 

\intitle{Qualitative analysis}
Table \ref{table:induced_cluster} shows four of the clusters created by our model and one cluster created by word2vec for the taxonomic categorization dataset.\footnote{See appendix for the full list of clusters of all clustering algorithms.}
The clusters formed by our algorithm are syntagmatic, associating words frequently observed together (e.g., tokens in cluster 1 are related to {\it snow activity}, while cluster 2 broadly relates to {\it water}). The cluster formed by word2vec embeddings is taxonomic (all tokens are food products).
Our results provide initial evidence that syntagmatic clusters emerge from an unsupervised training algorithm drawing on simple joint clustering of words and images.


\begin{table}[tb]
\centering
\begin{tabular}{p{0.92\columnwidth}} 
\toprule
{\bf Ours 1}
\sethlcolor{orange}\hl{skis}; axe; \sethlcolor{orange}\hl{sled}; \sethlcolor{red}\hl{parka}; \sethlcolor{orange}\hl{sleigh}; \sethlcolor{red}\hl{pants}; \sethlcolor{red}\hl{gloves} \\
 \midrule
{\bf Ours 2}
\sethlcolor{yellow}\hl{sailboat}; \sethlcolor{yellow}\hl{canoe}; \sethlcolor{green}\hl{swan}; \sethlcolor{yellow}\hl{raft}; \sethlcolor{yellow}\hl{boat}; \sethlcolor{yellow}\hl{yacht}; \sethlcolor{green}\hl{duck}; willow; \sethlcolor{yellow}\hl{ship}; drum \\
\midrule
{\bf Ours 3}
\sethlcolor{lightgray}\hl{train}; bullet; \sethlcolor{lightgray}\hl{subway}; tack; bridge; trolley \\
\midrule
{\bf Ours 4}
bedroom; \sethlcolor{lime}\hl{rocker}; \sethlcolor{pink}\hl{drapes}; \sethlcolor{lime}\hl{bed}; \sethlcolor{lime}\hl{dresser}; \sethlcolor{lime}\hl{sofa}; \sethlcolor{lime}\hl{couch}; piano; \sethlcolor{pink}\hl{curtains}; \sethlcolor{lime}\hl{cushion}; lamp; \sethlcolor{lime}\hl{chair}; fan; \sethlcolor{lime}\hl{bureau}; \sethlcolor{lime}\hl{stool}; cabin; book \\
\midrule
{\bf W2V cluster}
\sethlcolor{cyan}\hl{avocado}, \sethlcolor{cyan}\hl{walnut}, \sethlcolor{cyan}\hl{pineapple}, \sethlcolor{cyan}\hl{grapefruit}, \sethlcolor{cyan}\hl{coconut}, \sethlcolor{cyan}\hl{olive}, \sethlcolor{cyan}\hl{lime}, \sethlcolor{cyan}\hl{lemon}
\\
\bottomrule
\end{tabular}
\caption{Four clusters induced by our model (Ours 1, 2, 3, 4), sorted by $P(c|w)$, and one cluster induced by word2vec. Our clusters are syntagmatic, while the W2V cluster is taxonomic.
Words highlighted by the same color belong to the same taxonomic category.}
\label{table:induced_cluster}
\end{table}

\subsection{Concreteness Estimation} \label{concreteness_estimation}
\intitle{Background}
\citet{fisher_et_al_1994} suggest that the number of nouns in a sentence is among the earliest syntactic cues that children pick up.
Consequently, noun identification is assumed to be one of the first syntactic tasks learned by infants. We approximate noun identification as concreteness estimation, since words representing concrete entities are mostly nouns.\footnote{For example, in the concreteness dataset built by \citet{brysbaert_et_al_2013}, in which human annotators rated the concreteness of words on a scale of 1 to 5, 85.6\% of the words with an average concreteness rating above 4 are nouns.}
\citet{chang_2021} show that while children acquire concrete words first, neural text-based models show no such effect, suggesting that multimodality impacts the learning process.

\intitle{Setting}
We evaluate concreteness estimation using the dataset by \citet{brysbaert_et_al_2013}, which contains concreteness ratings for 40K English words averaged over multiple human annotated ratings on a scale of 1 to 5.
We estimate the concreteness of a word as the maximum probability with which it was assigned to any cluster.
For evaluation, we follow \citet{charbonnier_wartena_2019} and compute the Pearson correlation coefficient of our predictions with the ground-truth values. In addition, we investigate the impact of word frequency on our model's predictions by evaluating the model on subsets of words in the Brysbaert data of increasing minimum frequency in MSCOCO.

\intitle{Comparison with other models}
First, we compare against supervised {\bf SVM regression} models, which have shown strong performance on the Brysbaert data in prior work~\cite{charbonnier_wartena_2019}. Following their work, we use two feature configurations: (1)~POS tags + suffixes, (2)~POS tags + suffixes + pre-trained FastText embeddings~\cite{joulin_et_al_2017}. 
We train the SVMs on the full Brysbaert data. 

Second, we compare with a minimally supervised {\bf text-only} model. As in  Sec \ref{semantic_word_clustering}, we create word vector representations from co-occurrence counts.  Next, following prior work \cite{turney_2011}, we select concrete (abstract) representative words by taking the 20 words with the highest (lowest) concreteness value in the Brysbaert data that occur more than 10 times in the MSCOCO training set.
We predict a word's concreteness by computing its average cosine similarity to the concrete representative words minus the average of its cosine similarity to the abstract representative words.

\begin{figure}
    \centering
    \includegraphics[width=7.6cm]{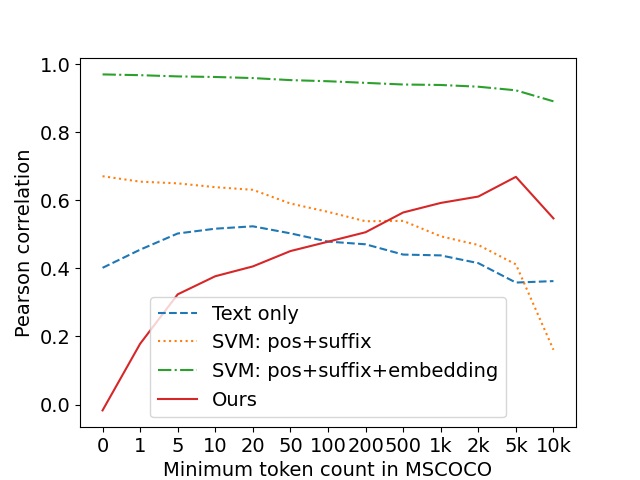}
    \caption{Pearson correlation between predicted word concreteness and gold standard human ratings, evaluated over test sets with increasing minimum frequency in the MSCOCO data. Results are averaged across 5 random initializations (std was consistently $<0.03$).}
    \label{fig:conc_with_baseline}
\end{figure}

\intitle{Results}
Figure \ref{fig:conc_with_baseline} presents the results in terms of Pearson correlation when evaluated on words of varying minimum frequency in MSCOCO. 
When considering frequent tokens only, our model predicts word concreteness with an accuracy higher than the SVM with POS and suffix features, although additional embedding features improve SVM performance further. 
Note that the supervised baseline was trained on the full data set, and hence evaluated on a subset of its training set. 
Our multimodal model performs better than its text-only variant for tokens that occur at least 100 times, even though the text-only model has received some supervision (by selecting the representative words).


\subsection{Visual Multi-Label Classification} \label{coco_multi_classification}

In addition to linguistic knowledge, infants acquire visual semantic knowledge with little explicit supervision, i.e., they learn to segment and classify objects.
To test whether our model also acquires such knowledge we evaluated it on the multi-label classification task: For each image in the MSCOCO test set, predict the classes of objects in the image.

In a zero-shot setting, we mapped the induced clusters to predicted lists of MSCOCO classes as follows.
We first provided the name of each class to our model as text
\ubst{input, retrieved the assigned cluster, and ran inference $C$ times, $C$ being the number of MSCOCO classes,}
input and retrieved the assigned cluster,
thus obtaining a (one-to-many) cluster-to-classes mapping. Now, for each test image, we used the visual encoder to predict the assigned cluster(s). The predicted set of MSCOCO classes is the union of the lists of classes to which the predicted clusters are mapped.


\intitle{Comparison with CLIP}
We compare our results against CLIP. 
To ensure comparability with our model we use CLIP with ResNet50.
We use CLIP as a point of comparison to provide perspective on the capabilities of our model despite differences in modeling and assumptions. However, we note two caveats regarding this comparison.
First, CLIP was trained on a much larger training set and has more parameters than our model (see  Table~\ref{table:models_statistics}). 
Second, CLIP has only been used for single- (not multi-) label classification, 
by inferring encodings of both input images and prompts representing the ground-truth classes (e.g., ``a photo of a bus'' for the ground truth class \emph{bus}) and assigning the image to the class with  highest cosine similarity to its encoding.
We adapt CLIP to a multi-label setting as follows: Instead of assigning the image to the class with the highest cosine similarity, we take into account the cosine similarity with all classes for each image. We consider a class as predicted if its cosine similarity exceeds a threshold,
tuned on the MSCOCO training split.

\intitle{Results}
Table \ref{table:classification_f_score} presents the results. As expected, CLIP outperforms our model. However, our model achieves impressive results considering its simplicity, its size, and that CLIP is the current state-of the-art in self-supervised vision and language learning. Training a CLIP model of comparable size and exposed to similar training data as our model is beyond the scope of this paper, but an interesting direction for future work. 

\begin{table}
\centering
\begin{tabular}{llll} 
\toprule
Model & Precision & Recall & F-Score \\
\midrule
Ours & \small{0.43 $\pm$ 0.04} & \small{0.21 $\pm$ 0.02} & \small{0.28 $\pm$ 0.01} \\
CLIP & \small{0.52} & \small{0.39} & \small{0.45} \\
\bottomrule
\end{tabular}
\caption{Visual multi-label classification results on the MSCOCO data for our model and CLIP in terms of precision, recall and F1 score. We report mean (std) over 5 random initializations for our model.
CLIP experiments are deterministic (we use the pretrained model directly, unlike Sec \ref{semantic_word_clustering} where we used KMeans on top of CLIP).}
\label{table:classification_f_score}
\end{table}

\subsection{Object Localization} \label{object_localization}
Another important task performed by infants is visual object localization.
To test our model's ability to reliably localize objects in images we use Class Activation Maps (CAM) described by \citet{zhou_et_al_2016}. Each CAM indicates how important each pixel was during classification for a specific cluster.

\intitle{Quantitative analysis}
Most previous studies of zero-shot segmentation \cite{bucher2019zero} trained on a subset of ``seen'' classes, and evaluated on both seen and unseen classes. We use a more challenging setup previously referred to as annotation-free segmentation \cite{zhou2021denseclip}, where we evaluate our model without any training for the segmentation task.
We use MSCOCO's ground-truth bounding boxes, which are human annotated and mark objects in the image, for evaluation.
Following the original CAM paper, we use a heuristic method to predict bounding boxes: Given a CAM, we segment the pixels of which the value is above 50\% of the max value of the CAM and take the bounding box that covers the largest connected component in the segmentation map.

We use precision and recall for evaluation. A pair of bounding boxes is considered a match if the intersection over union (IoU) of the pair exceeds 0.5.
Given lists of predicted and ground-truth bounding boxes, we consider each matched pair as a true positive and a prediction (ground-truth) for which no matching ground-truth (prediction) was found as a false positive (negative).
We compare our model to a random baseline: Sample $k$ random bounding boxes (where $k$ is the number of ground-truth bounding boxes in the current image).
This baseline uses the number of ground-truth bounding boxes in each image (our model is not exposed to this information).

The results are presented in Table \ref{table:heatmap_f_score}. Our model is significantly more precise than the random baseline, but achieves similar recall: the entire MSCOCO test split contains a total of 164,750 bounding boxes, while our model predicted 38,237 bounding boxes. This problem could be addressed by lowering the visual threshold. We leave this direction for future research.

\begin{table}[tb]
\centering
\begin{small}
\begin{tabular}{llll} 
\hline
Model & Precision & Recall & F-Score \\
\hline
Ours & \bf{0.178 $\pm$ 0.01} & 0.025 $\pm$ 0.004 & 0.044 $\pm$ 0.006 \\
Rand & 0.027 $\pm$ 0.001 & 0.027 $\pm$ 0.001 & 0.027 $\pm$ 0.001 \\
\hline
\end{tabular}
\end{small}
\caption{Mean F-score and standard deviation across 5 random initializations on bounding box prediction.
Our model (Ours) improves precision (bold) over random (Rand) significantly while achieving similar recall (might improve by tuning the visual threshold).}
\label{table:heatmap_f_score}
\end{table}

\begin{figure}[tb]
\centering
\includegraphics[width=5cm]{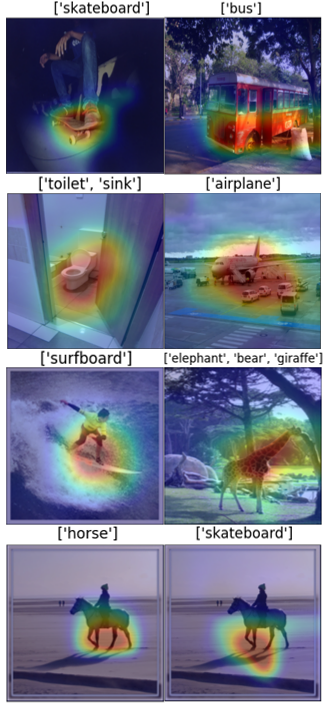}
\caption{Examples of heatmaps of CAM values. Above each image we print the class predicted by the model.}
\label{fig:all_heatmap}
\end{figure}

\intitle {Qualitative analysis}
Fig.~\ref{fig:all_heatmap} shows a selection of CAMs, plotted as heatmaps and associated with class predictions (see Sec.~\ref{coco_multi_classification}).
The heatmaps extracted by the model were better when the model predicted a correct class in the visual classification task (top six images and bottom left image in Fig \ref{fig:all_heatmap}).
In the bottom two images two clusters were predicted for the same original image, one correct and one incorrect (with an, unsurprisigly, meaningless heatmap).

\section{Discussion and Conclusion} \label{conclusion}

We proposed a model for unsupervised multimodal lagnguage acquisition, trained to jointly cluster text and images.
%
Many of our design choices were guided by findings from cognitive studies of infant language acquisition: The joint learning of multiple modalities; learning word-level semantics \citep[e.g.,][suggest that children first learn to identify nouns and use this information to learn sentence-level semantics]{fisher_et_al_1994}; and cross-situational learning \citep[counting how many times each word co-occurred with each cluster, see][]{gleitman_1990}. After training, our model demonstrates capabilities typical of infant language acquisition: Word concreteness prediction and identification and segmentation of objects in a visual scene.

However, we do not stipulate that infants begin their acquisition of language  by clustering words. It would be interesting to design experiments to test this hypothesis, e.g., by connecting our work with laboratory work on joint word and category learning~\cite{borovsky2006language}, or work on the emergence of syntagmatic vs. taxonomic categories in young children~\cite{sloutsky_2017}.

While our model is cognitively more plausible compared to previous studies, the gap from a realistic setting of language acquisition is still large: (1) we assume the language input is segmented into words; (2) the input data, while naturalistic, is not typical of infants at the stage of language acquisition; (3) the input only includes the visual and textual modality, but not, e.g., pragmatic cues like gesture; and (4) the model learns in a non-interactive setting, whereas physical and social interactions are considered crucial for language learning, and learns (and is evaluated) in a batch fashion while human learning is typically incremental~\cite{frermann_lapata_2016}.

In the semantic word categorization and concreteness prediction experiments, we compared our multimodal model to unimodal text-only baselines, which we chose to be as similar as possible to our model. The results suggest that multimodality improves performance on both text tasks. However, it is unclear which specific information is encoded in the visual modality that benefits these text tasks. We leave this question for future research.

Syntagmatic categories, although highly intuitive in the context of human memory, were not the subject of many previous computational studies. We propose to further investigate this type of categories and its use.
One interesting direction is to combine syntagmatic categories with interactivity: Given a relevant signal from the environment the model can cycle through concepts in the syntagmatic category triggered by the signal, speeding up the extraction of relevant concepts in real time. One possible application of this direction is modelling the construction of ad-hoc categories, described by \citet{barsalou_1983}.

While all experiments are conducted in English, our setting supports future work on other languages. The small training set and the nature of the data (image-sentence pairs that might be, to some extent, collected from the Internet) allow our model to be extended to low-resource languages, while the minimal structural assumptions of the text encoder may imply some degree of language-agnosticity.

In future work, we plan to improve the cognitive plausibility of our model by
(1) incorporating typical input observed by children \citep[by using videos taken in real scenes of child language acquisition, see][]{sullivan_et_al_2021}; and (2) changing the setting to an interactive one, where the model is transformed into a goal-driven agent that uses interactivity to learn and produce language.


\section*{Acknowledgements}
We would like to thank the anonymous reviewers for their helpful comments and feedback.
This work was supported in part by the Israel Science Foundation (grant no. 2424/21), by  a  research gift from the Allen Institute for AI and by the HUJI-UoM joint PhD program.

\section*{Ethical Considerations}
We used publicly available resources to train our model ~\citep{lin_et_al_2014}. As with other statistical methods for word representations, our approach may capture social biases which manifest in its training data \citep[e.g., MSCOCO was shown to be biased with respect to gender,][]{zhao2017men}. Our code includes a model card~\citep{mitchell2019model} which reports standard information regarding the training used to produce our models and its word clusters.

\bibliography{bib}

\appendix
\section{Appendix}
\label{sec:appendix}

\subsection{Implementation details}
\label{app:implementation}
\subsubsection{Training}

For the visual encoder, we used the ResNet50 implementation from the torchvision package with ADAM optimizer and a learning rate of $10^{-4}$.

\subsubsection{Semantic Word Categorization}

For clustering of word embeddings, we use the K-Means implementation in scikit-learn.
We use the word2vec google-news-300 model from the gensim package, the $\text{BERT}_ \text{BASE}$ model from the transformers library and CLIP's official implementation.

\subsubsection{Concreteness Estimation}

\intitle{Supervised model}
For the implementation of the supervised model described by \cite{charbonnier_wartena_2019} we used 3 feature types. For POS tag, we used the WordNet module from the NLTK package to find, for each word, the number of synsets in each of the 4 possible POS tags (NOUN, VERB, ADJ, ADV), and the induced feature vector was the normalized count vector. In case no synsets were found the induced feature vector was all zeros. For suffixes, we collected the 200 most frequent suffixes of length 1 to 4 characters in the training set, and the induced feature vector was a 200-d binary vector indicating, for each suffix, if it occurs in the current word. For word embeddings, we used the fastText wiki-news-300d-1M model. The final feature vector is the concatenation of all the selected feature vectors: POS+suffix for the first model (resulting a 204-d feature vector for each word), and POS+suffix+embedding for the second model (resulting a 504-d feature vector for each word).

\intitle{Text-only baseline}
The 20 concrete representative words were \emph{sand}, \emph{seagull}, \emph{snake}, \emph{snowsuit}, \emph{spaghetti}, \emph{stairs}, \emph{strawberry}, \emph{tiger}, \emph{tomato}, \emph{toothbrush}, \emph{tractor}, \emph{tree}, \emph{turtle}, \emph{umbrella}, \emph{vase}, \emph{water}, \emph{comb}, \emph{tire}, \emph{firetruck}, \emph{tv}.

The 20 abstract representative words were \emph{would}, \emph{if}, \emph{though}, \emph{because}, \emph{somewhat}, \emph{enough}, \emph{as}, \emph{could}, \emph{how}, \emph{yet}, \emph{normal}, \emph{ago}, \emph{so}, \emph{very}, \emph{the}, \emph{really}, \emph{then}, \emph{abstract}, \emph{a}, \emph{an}.

\subsubsection{Object Localization}

To extract Class Activation Mappings, we used the CAM module from the torchcam package.

\subsection{Cluster lists}

Following is a list of clusters created by different clustering algorithm, in a specific execution.

\subsubsection{Our model}

Words are sorted by $P(c|w)$.
\\~\\
\setlength{\parindent}{0pt}
\textbf{Cluster 1}: doorknob, canary

\textbf{Cluster 2}: trombone, leotards, trumpet, projector, cello, harmonica, guitar

\textbf{Cluster 3}: train, bullet, subway, tack, bridge, trolley

\textbf{Cluster 4}: bus, ambulance, inn, taxi, level

\textbf{Cluster 5}: elephant, bear, giraffe, paintbrush, rock, fence, chain

\textbf{Cluster 6}: machete, porcupine, hornet, banana, gorilla, apple, turtle, turnip, peach, stick

\textbf{Cluster 7}: veil, shawl

\textbf{Cluster 8}: ashtray, mushroom, cheese, spinach, olive, tomato, shrimp, rice, pie, chicken, potato, broccoli, plate, pan, pepper, asparagus, skillet, peas, onions, tuna, salmon, cranberry, lettuce, beans, spatula, ladle, dish, crab, corn, cucumber, tray, walnut, plum, box, lobster, cherry, table, shell

\textbf{Cluster 9}: church, clock, skyscraper, chapel, building, brick, stone, flea

\textbf{Cluster 10}: airplane, helicopter, pier, gate

\textbf{Cluster 11}: bedroom, rocker, drapes, bed, dresser, sofa, couch, piano, curtains, cushion, lamp, chair, fan, bureau, stool, cabin, book

\textbf{Cluster 12}: skis, axe, sled, parka, sleigh, pants, gloves

\textbf{Cluster 13}: dishwasher, kettle, toaster, freezer, stove, microscope, microwave, oven, fridge, cupboard, mixer, blender, plug, mittens, grater, pot, apron, cabinet, tape, apartment

\textbf{Cluster 14}: missile, jet, bomb, rocket, drill

\textbf{Cluster 15}: bouquet, thimble, umbrella, accordion, cake, scissors, wrench, jar, pliers, candle, penguin, frog, doll, bottle, shield, pig, card

\textbf{Cluster 16}: zucchini, beets, cabbage, celery, cauliflower, wheelbarrow, parsley, tongs, shelves

\textbf{Cluster 17}: grapefruit, tangerine, colander, clamp, snail, cantaloupe, pineapple, grape, pear, lemon, eggplant, mandarin, garlic, nectarine, basket, corkscrew, pyramid, pumpkin, bin, sack, lime, cork, orange

\textbf{Cluster 18}: octopus, kite, crocodile, squid, balloon, butterfly, whale

\textbf{Cluster 19}: surfboard, swimsuit, board, rope

\textbf{Cluster 20}: hose, hut

\textbf{Cluster 21}: skateboard, pipe, saxophone, helmet, escalator, barrel, broom

\textbf{Cluster 22}: shotgun, seal, dolphin, car, hoe, hamster, wheel, house

\textbf{Cluster 23}: sailboat, canoe, swan, raft, boat, yacht, duck, willow, ship, drum

\textbf{Cluster 24}: tortoise, dog, cat, tiger, cheetah

\textbf{Cluster 25}: hyena

\textbf{Cluster 26}: buckle, mug, ruler, envelope, bag, belt, cup, camel, pencil, spider, cart, saucer, closet, tripod, carpet

\textbf{Cluster 27}: crowbar, bathtub, toilet, drain, sink, faucet, marble, mirror, basement, tank, bucket, door, razor, mat

\textbf{Cluster 28}: toad, mouse, keyboard, key, desk, typewriter, stereo, rat, bookcase, telephone, anchor, radio

\textbf{Cluster 29}: buzzard, chickadee, finch, woodpecker, grasshopper, worm, sparrow, blackbird, vulture, parakeet, bluejay, hawk, robin, dagger, perch, falcon, stork, peacock, pelican, owl, crow, pigeon, seagull, flamingo, eagle, vine, birch, beaver, pheasant, raven, goose, squirrel, seaweed, ant, emu, dove, cage, crown, shovel

\textbf{Cluster 30}: horse, racquet, saddle, pony, buggy, bat, football, wagon, sword, donkey, ball, fox

\textbf{Cluster 31}: beetle

\textbf{Cluster 32}: zebra, ostrich, elk, deer, lion, pen, pin, rifle, bolts

\textbf{Cluster 33}: bracelet, fawn, slippers, socks, shoes, tap, boots, strainer, jeans, ring

\textbf{Cluster 34}: whistle, cathedral, wand, thermometer, peg, hook, goldfish, lantern, wall, urn, caterpillar, chandelier, robe, leopard

\textbf{Cluster 35}: motorcycle, bike, tractor, truck, trailer, tricycle, scooter, jeep, limousine, garage, van, tent, crane

\textbf{Cluster 36}: baton, revolver, violin, tie, bow, cockroach, elevator, mink, necklace, blouse, trousers, vest, scarf, skirt, gun, gown, dress, shirt, sweater, bra, cap, jacket, coat, cape

\textbf{Cluster 37}: bench, cannon

\textbf{Cluster 38}: unicycle, groundhog

\textbf{Cluster 39}: pistol, buffalo

\textbf{Cluster 40}: clam, pickle, raisin, raspberry, napkin, submarine, fork, coconut, strawberry, bread, spoon, blueberry, radish, knife, biscuit, cloak, spear, whip, avocado, carrot, cottage, turkey, bowl

\textbf{Cluster 41}: lamb, sheep, raccoon, cow, goat, rooster, calf, ox, hatchet, bull, moose, bison, barn, rabbit, shed, shack

\textbf{Cluster 42}: screwdriver, pajamas, comb, hammer, brush, alligator

\subsubsection{word2vec}

\textbf{Cluster 1}: leopard, hyena, crocodile, canary, lion

\textbf{Cluster 2}: lobster, tuna, clam, octopus, whale, squid, shrimp, seaweed, salmon, crab, dolphin

\textbf{Cluster 3}: mat, cage

\textbf{Cluster 4}: lantern, chandelier, candle, tripod, projector, lamp

\textbf{Cluster 5}: sailboat, submarine, raft, yacht, canoe, boat, pier, ship

\textbf{Cluster 6}: avocado, walnut, pineapple, grapefruit, coconut, olive, lime, lemon

\textbf{Cluster 7}: mittens, doll, slippers, pajamas, necklace, socks

\textbf{Cluster 8}: rock, cottage, tent, gate, house, brick, pyramid, rocker, door, bluejay, shed, bench, skyscraper, bolts, hut, mirror, key, building, barrel, tape, inn, apartment, cabinet, book, marble, drum, shack, umbrella, crane, bureau, garage, shell, basement, fan, cathedral, fence, chapel, stone, drill, telephone, comb, radio, shield, church, anchor, microscope, clock, level, board, football, chain, cabin, wall, barn, bridge

\textbf{Cluster 9}: elk, bison, pheasant, beaver, deer, moose, goose

\textbf{Cluster 10}: pig, cow, sheep, goat, ostrich, emu, calf, buffalo, bull

\textbf{Cluster 11}: elevator, train, whistle, limousine, escalator, subway, bus, taxi, trolley

\textbf{Cluster 12}: groundhog, parakeet, fawn, tortoise, goldfish, porcupine, fox, cheetah, gorilla, flea, rabbit, mink, peacock, rooster, mouse, duck, turtle, squirrel, dog, bear, alligator, rat, raccoon, cat, flamingo, tiger, hamster, penguin

\textbf{Cluster 13}: razor, pliers, scissors, crowbar, knife, screwdriver, machete

\textbf{Cluster 14}: keyboard, violin, trumpet, piano, saxophone, guitar, cello, accordion, trombone, harmonica

\textbf{Cluster 15}: rocket, helicopter, jet, bomb, missile, ambulance, airplane

\textbf{Cluster 16}: jeans, leotards, boots, blouse, skirt, bracelet, shirt, swimsuit, shoes, trousers, dress, pants, sweater, bra

\textbf{Cluster 17}: cauliflower, spinach, cabbage, broccoli, peas, garlic, radish, lettuce, eggplant, cucumber, onions, zucchini, parsley, celery, beans, asparagus, beets

\textbf{Cluster 18}: sofa, drapes, typewriter, napkin, toilet, chair, bathtub, bedroom, bed, doorknob, stool, desk, carpet, table, dresser, couch, stereo, curtains

\textbf{Cluster 19}: strainer, colander

\textbf{Cluster 20}: kite, balloon, willow

\textbf{Cluster 21}: corn, pickle, bread, turkey, biscuit, dish, cheese, cake, lamb, pepper, pie, rice, chicken

\textbf{Cluster 22}: frog, spider, toad, ant, worm, cockroach, snail, butterfly, beetle, hornet, grasshopper, caterpillar

\textbf{Cluster 23}: cannon, bullet, gun, pistol, rifle, revolver, shotgun

\textbf{Cluster 24}: bag, kettle, mug, envelope, sack, urn, basket, cup, pot, box, card, plate, jar, bucket, bouquet, bin, ashtray, tray, bottle

\textbf{Cluster 25}: bowl, spatula, spoon, blender, ladle, grater, tongs, pan, mixer, saucer

\textbf{Cluster 26}: sleigh, trailer, buggy, wheelbarrow, van, wagon, tractor, truck, cart, jeep

\textbf{Cluster 27}: plum, cherry, birch, cork

\textbf{Cluster 28}: finch, falcon, pigeon, hawk, pelican, raven, seagull, stork, buzzard, vulture, chickadee, sparrow, robin, crow, owl, woodpecker, blackbird, eagle, swan

\textbf{Cluster 29}: racquet, bike, skateboard, skis, unicycle, sled, scooter, motorcycle, helmet, surfboard, wheel, saddle, tricycle, car

\textbf{Cluster 30}: sword, ruler, dagger, spear, baton

\textbf{Cluster 31}: bookcase, shelves, closet, fridge, cupboard

\textbf{Cluster 32}: thermometer, microwave, dishwasher, toaster, skillet, stove, oven, freezer

\textbf{Cluster 33}: vest, coat, jacket, parka, gloves

\textbf{Cluster 34}: plug, thimble, tap, seal, dove, sink, drain

\textbf{Cluster 35}: shawl, scarf, cap, cloak, veil, gown, cape, robe, wand, apron

\textbf{Cluster 36}: hammer, broom, shovel, pencil, hatchet, brush, paintbrush, hoe, wrench, bat, pen

\textbf{Cluster 37}: clamp

\textbf{Cluster 38}: pumpkin, vine, grape, raspberry, carrot, mandarin, strawberry, pear, banana, apple, turnip, nectarine, cantaloupe, orange, mushroom, peach, cranberry, tomato, tangerine, raisin, blueberry, potato

\textbf{Cluster 39}: faucet, tank, pipe, hose

\textbf{Cluster 40}: donkey, ox, pony, horse, camel, elephant, zebra, giraffe

\textbf{Cluster 41}: bow, belt, tie, stick, buckle, cushion, hook, peg, perch, ring, tack, pin, ball, corkscrew, fork, whip, rope, crown

\subsubsection{BERT}

\textbf{Cluster 1}: crane, vulture, finch, pigeon, owl, sparrow, snail, octopus, bat, lobster, crab, mushroom, shrimp, shell, squid, perch, hornet, spider, worm, butterfly, turtle, toad

\textbf{Cluster 2}: anchor, tack, bow, raft, doll, tray, knife, jet, airplane, canoe, car, helicopter, boat, ship, napkin, book, board, card, desk, chair, bed, sword, bomb, dagger, spear, rope, bag, pencil

\textbf{Cluster 3}: pheasant, woodpecker, parakeet, ostrich, caterpillar

\textbf{Cluster 4}: stork, fawn, hatchet, hyena, raccoon, grasshopper

\textbf{Cluster 5}: pliers, toaster, mittens, strainer, blender, freezer, saucer

\textbf{Cluster 6}: crow, eagle, raven, hawk, dove, pig, sheep, fox, dog, cat, peacock, camel, bear, elephant, deer, buffalo, rabbit, dolphin, frog, cow, elk, lion, moose, donkey, beaver, squirrel, rat, mouse, salmon, goat, calf, whale, leopard, bison, horse, bull, crocodile

\textbf{Cluster 7}: hook, tape, pipe, pyramid, mat, chain, drill, balloon, ball, kite, cap, ring, belt, umbrella, bin, bucket, barrel, basket, bench, gate, wheel, plug, key, stereo, mixer, baton, envelope

\textbf{Cluster 8}: scooter, sleigh, shawl, sled

\textbf{Cluster 9}: raisin, raspberry, beets

\textbf{Cluster 10}: birch, grape, pear, plum, apple, cherry, orange, tomato, peach, lemon, peas, beans, pepper, carrot, lime, rice, potato, olive, garlic, corn, walnut, strawberry, cheese, coconut, mandarin, cabbage, banana, vine, willow, onions

\textbf{Cluster 11}: pelican, chickadee, porcupine, cucumber, cockroach, tortoise

\textbf{Cluster 12}: trailer, hose, saddle, tractor, ambulance, wagon, taxi, bus, submarine, subway, train, elevator, limousine, bike, trolley, motorcycle, jeep, truck, yacht, tank, sofa, rocket, missile, cart, helmet

\textbf{Cluster 13}: skillet, ladle

\textbf{Cluster 14}: level, building, bridge, pier, house, cabin, shield, lantern, marble, sink, apartment, hut, basement, wall, cottage, box, rock, door, table, cage, fence, brick, lamp, telephone, drain, shed, garage, stone, skyscraper, barn, church, cathedral, chapel

\textbf{Cluster 15}: buggy

\textbf{Cluster 16}: revolver, rifle, pistol, shotgun, cannon, gun, bullet

\textbf{Cluster 17}: seagull, sailboat, seaweed

\textbf{Cluster 18}: urn

\textbf{Cluster 19}: wheelbarrow, doorknob

\textbf{Cluster 20}: ruler, shovel, stove, keyboard, microscope, colander, cupboard, bowl, dish, skis, tie, pie, bread, cake, toilet, stool, cushion, mirror, tap, cabinet, carpet, fork, comb, apron

\textbf{Cluster 21}: skateboard, surfboard, swimsuit

\textbf{Cluster 22}: bluejay, blackbird, nectarine, grapefruit, tangerine, eggplant, asparagus, cauliflower, pineapple, cranberry, blueberry, goldfish, groundhog, mink, broccoli

\textbf{Cluster 23}: flamingo, hoe, racquet, parka

\textbf{Cluster 24}: violin, accordion, piano, cello, guitar, trombone, harmonica, trumpet, saxophone

\textbf{Cluster 25}: chandelier

\textbf{Cluster 26}: buzzard

\textbf{Cluster 27}: hamster

\textbf{Cluster 28}: gown, robe, bra, scarf, sweater, shirt, jacket, skirt, coat, dress, necklace, bouquet, blouse

\textbf{Cluster 29}: parsley, biscuit, celery

\textbf{Cluster 30}: goose, duck, falcon, rooster, canary, turkey, swan, gorilla, penguin, tiger, zebra, fan, pajamas, pants, jeans, van, pumpkin, tuna, chicken, rocker, lamb, ox, pony, ant, flea, beetle, cape, alligator

\textbf{Cluster 31}: thermometer

\textbf{Cluster 32}: machete

\textbf{Cluster 33}: cheetah, giraffe

\textbf{Cluster 34}: wrench, corkscrew, screwdriver, escalator, tongs

\textbf{Cluster 35}: thimble, tricycle, unicycle, tripod

\textbf{Cluster 36}: avocado, lettuce

\textbf{Cluster 37}: robin, hammer, pin, pen, bolts, scissors, brush, microwave, fridge, oven, drum, bedroom, curtains, peg, football, wand, mug, pot, shoes, trousers, vest, cloak, socks, boots, tent, inn, gloves, razor, bracelet, crown, buckle, shack, bottle, sack, plate, broom, candle, cork, dresser, couch, bureau, seal, whip, cup, clock, radio, closet, jar, shelves, kettle, pan, stick, spoon, veil, whistle

\textbf{Cluster 38}: crowbar, emu, clam, drapes, zucchini, radish, turnip, clamp, projector, bookcase, spatula, grater, spinach, pickle

\textbf{Cluster 39}: paintbrush, typewriter

\textbf{Cluster 40}: cantaloupe, leotards

\textbf{Cluster 41}: dishwasher, faucet, slippers, ashtray, bathtub

\subsubsection{CLIP}

\textbf{Cluster 1}: mittens, doll, rabbit, mouse, squirrel, rat, cat, hamster

\textbf{Cluster 2}: plug, lantern, kettle, mug, thimble, urn, cup, candle, pot, blender, jar, book, bucket, toaster, bin, bottle, cage, lamp

\textbf{Cluster 3}: mirror, ashtray, table, tray, mat

\textbf{Cluster 4}: chandelier, bracelet, basket, unicycle, bolts, cap, barrel, tape, drum, umbrella, shell, necklace, stool, bouquet, ring, fan, tack, drill, telephone, wheel, saddle, microscope, clock, whip, chain, rope, crown, hose

\textbf{Cluster 5}: bow, broom, shovel, spatula, spoon, ladle, tongs, crowbar, spear, fork

\textbf{Cluster 6}: keyboard, typewriter, raft, piano, escalator, comb, sink, drain, accordion

\textbf{Cluster 7}: pumpkin, bread, biscuit, worm, carrot, cheese, orange, tangerine

\textbf{Cluster 8}: trailer, rocker, bike, gun, box, train, motorcycle, projector, van, tractor, radio, bus, truck, mixer, taxi, tank, car, ambulance, jeep

\textbf{Cluster 9}: jeans, leotards, bag, blouse, skirt, sack, tie, shirt, swimsuit, trousers, pajamas, socks, dress, carpet, veil, gown, pants, sweater, curtains, apron

\textbf{Cluster 10}: peacock, raven, robin, crow, blackbird

\textbf{Cluster 11}: elk, fawn, bison, cheetah, leopard, deer, emu, hyena, moose, zebra, giraffe

\textbf{Cluster 12}: donkey, ox, cow, pony, horse, camel, sheep, goat, lamb, calf, buffalo, bull

\textbf{Cluster 13}: thermometer, bullet, stick, tripod, pencil, ruler, peg, brush, paintbrush, hoe, pin, screwdriver, baton, wand, pen

\textbf{Cluster 14}: pepper, beans

\textbf{Cluster 15}: racquet, skateboard, skis, scooter, doorknob, surfboard, guitar, board

\textbf{Cluster 16}: plum, vine, grape, raspberry, cherry, olive, cranberry, raisin, blueberry

\textbf{Cluster 17}: pig, groundhog, porcupine, fox, seal, gorilla, beaver, whale, dog, bear, elephant, raccoon, salmon, lion, tiger

\textbf{Cluster 18}: violin, trumpet, saxophone, cello, trombone, harmonica

\textbf{Cluster 19}: parakeet, bluejay, finch, mink, dove, perch, birch, canary, bat, chickadee, sparrow, woodpecker

\textbf{Cluster 20}: sleigh, cannon, buggy, sled, canoe, wheelbarrow, limousine, wagon, tricycle, cart, trolley

\textbf{Cluster 21}: shed, elevator, garage, basement, barn

\textbf{Cluster 22}: avocado, walnut, pineapple, grapefruit, coconut, marble, strawberry, pear, apple, lime, nectarine, cantaloupe, peach, willow, tomato, lemon, potato

\textbf{Cluster 23}: crane, ostrich, pelican, stork, flamingo

\textbf{Cluster 24}: rock, boots, brick, slippers, shoes, helmet, stone, ball, bomb, balloon, football, bra

\textbf{Cluster 25}: flea

\textbf{Cluster 26}: radish, turnip, parsley, beets

\textbf{Cluster 27}: bookcase, shelves, cabinet, bureau, closet, desk, dresser, fridge, freezer, stereo, cupboard

\textbf{Cluster 28}: cottage, tent, house, pyramid, door, skyscraper, card, hut, bedroom, building, inn, apartment, shack, cathedral, chapel, church, level, cabin, wall

\textbf{Cluster 29}: lobster, spider, octopus, ant, squid, shrimp, cockroach, beetle, hornet, crab, grasshopper

\textbf{Cluster 30}: tuna, tortoise, frog, goldfish, toad, clam, turtle, mandarin, snail, butterfly, alligator, crocodile, mushroom, dolphin

\textbf{Cluster 31}: sailboat, submarine, yacht, boat, rocket, helicopter, jet, missile, ship, airplane

\textbf{Cluster 32}: envelope, bowl, napkin, toilet, bathtub, microwave, plate, dishwasher, dish, cake, skillet, stove, pan, shield, pie, oven, rice, saucer

\textbf{Cluster 33}: shawl, vest, scarf, coat, drapes, cloak, jacket, parka, cape, robe

\textbf{Cluster 34}: hammer, belt, razor, sword, pliers, tap, key, hatchet, buckle, hook, whistle, pistol, rifle, revolver, kite, faucet, clamp, scissors, wrench, gloves, dagger, knife, anchor, corkscrew, shotgun, machete, pipe, cork

\textbf{Cluster 35}: cauliflower, cabbage, garlic, onions

\textbf{Cluster 36}: turkey, pheasant, falcon, pigeon, hawk, rooster, duck, seagull, buzzard, vulture, chicken, owl, goose, eagle, penguin, swan

\textbf{Cluster 37}: gate, pier, fence, subway, bridge

\textbf{Cluster 38}: spinach, broccoli, lettuce, seaweed

\textbf{Cluster 39}: sofa, bench, chair, bed, cushion, couch

\textbf{Cluster 40}: corn, pickle, peas, eggplant, cucumber, banana, zucchini, celery, asparagus, caterpillar

\textbf{Cluster 41}: strainer, grater, colander

\subsubsection{Text-only}

\textbf{Cluster 1}: saxophone, buckle, broom, shotgun, hatchet

\textbf{Cluster 2}: grape, pepper, potato, lettuce

\textbf{Cluster 3}: pipe, racquet, skateboard, tricycle, skis, barrel, board, helmet

\textbf{Cluster 4}: emu, mug, cup

\textbf{Cluster 5}: eagle, ostrich, thermometer, owl, elephant, octopus, accordion, apple, orange, jet, airplane, apartment, umbrella, ox, escalator

\textbf{Cluster 6}: surfboard, pajamas, swimsuit

\textbf{Cluster 7}: falcon, crow, pigeon, bluejay, raven, hawk, tack, snail, blackbird, mat, peg, tray, cloak, submarine, limousine, belt, radish, lobster, biscuit, coconut, turnip, napkin, stool, sofa, cushion, couch, bench, table, squirrel, perch, seal, vine, pan, saucer, envelope, carpet

\textbf{Cluster 8}: cantaloupe, zucchini, parsley, eggplant, pineapple, beets, garlic, cranberry, mandarin, celery

\textbf{Cluster 9}: hammer, shovel, crowbar, screwdriver, dolphin, drill, hose, bat, wand, rifle, sword, bomb, cockroach, clamp, gun

\textbf{Cluster 10}: wrench, thimble, scissors, pliers, nectarine, spear

\textbf{Cluster 11}: robin, level, pier, trailer, shield, tractor, ambulance, taxi, bus, bike, trolley, car, motorcycle, jeep, truck, inn, beetle, garage

\textbf{Cluster 12}: bolts

\textbf{Cluster 13}: flamingo, wheelbarrow, machete, porcupine, harmonica, hut, tuna, bin, goldfish, hyena, ant, willow, bouquet, strainer, missile, tongs

\textbf{Cluster 14}: birch, vulture, finch, pin, pen, brush, sheep, bear, deer, buffalo, zebra, elk, pyramid, cabin, basement, cage, lamb, calf, bison, giraffe, flea, grasshopper, barn

\textbf{Cluster 15}: pheasant, dove, ruler, paintbrush, fan, chandelier, piano, drum, cherry, drapes, bedroom, curtains, razor, peach, dresser, bureau, rocker, lamp, radio, telephone, closet, bookcase, shelves, grater, comb

\textbf{Cluster 16}: frog, keyboard, desk, mouse, key

\textbf{Cluster 17}: asparagus, cauliflower, lemon, peas, beans, rice, corn, chicken, shrimp, cabbage, salmon, broccoli

\textbf{Cluster 18}: goose, duck, seagull, woodpecker, swan, anchor, fox, sparrow, moose, sailboat, boat, ship, yacht, urn, gate, rocket, cannon, cathedral

\textbf{Cluster 19}: raccoon

\textbf{Cluster 20}: crane, rooster, stork, parakeet, pelican, hook, hoe, pig, cheetah, gorilla, dog, peacock, camel, rabbit, penguin, cow, lion, donkey, fridge, toaster, guitar, trombone, building, bridge, house, chain, lantern, football, raft, scooter, balloon, ball, kite, doll, robe, bra, scarf, socks, van, canoe, helicopter, tent, necklace, bracelet, ring, crown, carrot, banana, shack, wall, bottle, bucket, sack, tank, box, basket, rock, door, book, card, candle, cork, chair, bed, fence, brick, rat, goat, pony, whale, leopard, bull, mink, spider, worm, tap, rope, wheel, clock, projector, blender, tripod, drain, typewriter, mixer, cart, jar, shed, bag, freezer, sled, stone, stick, spatula, ladle, butterfly, alligator, turtle, skyscraper, church

\textbf{Cluster 21}: ashtray

\textbf{Cluster 22}: revolver

\textbf{Cluster 23}: mittens, tomato, olive, mushroom, cheese, crocodile, spinach, onions

\textbf{Cluster 24}: chapel

\textbf{Cluster 25}: clam

\textbf{Cluster 26}: baton

\textbf{Cluster 27}: canary, cat, fawn, tiger, shoes, slippers, hamster, beaver, groundhog

\textbf{Cluster 28}: unicycle

\textbf{Cluster 29}: subway, train, bullet

\textbf{Cluster 30}: turkey, violin, cello, bow, cap, gown, trousers, parka, sweater, shirt, jacket, skirt, pants, shawl, coat, vest, jeans, dress, boots, elevator, gloves, tie, pistol, blouse, apron, veil, cape

\textbf{Cluster 31}: pear, raisin, cucumber, avocado, hornet, plug

\textbf{Cluster 32}: tape, grapefruit, tangerine, plum, raspberry, microscope, colander, bowl, knife, dish, pumpkin, crab, lime, pie, walnut, strawberry, bread, cake, blueberry, cottage, plate, shell, squid, caterpillar, seaweed, whip, pencil, fork, spoon, pickle

\textbf{Cluster 33}: toad

\textbf{Cluster 34}: saddle, wagon, buggy, sleigh, horse

\textbf{Cluster 35}: corkscrew, microwave, stove, oven, pot, stereo, skillet, kettle

\textbf{Cluster 36}: doorknob, marble, dishwasher, faucet, cupboard, sink, toilet, mirror, bathtub, cabinet

\textbf{Cluster 37}: leotards

\textbf{Cluster 38}: buzzard, chickadee

\textbf{Cluster 39}: trumpet

\textbf{Cluster 40}: whistle

\textbf{Cluster 41}: dagger, tortoise

\end{document}